\documentclass[sigconf]{acmart}
\AtBeginDocument{%
  }

\setcopyright{acmlicensed}
\copyrightyear{2026}
\acmYear{2026}
\acmDOI{XXXXXXX.XXXXXXX}
\usepackage{pgfplots}
\pgfplotsset{compat=1.18}
\usepackage{pdflscape}

\usepackage{ragged2e}
\usepackage{tabularx}
\usepackage{multirow}
\newcolumntype{L}{X}
\newcolumntype{C}{>{\centering\arraybackslash}X}
\newcolumntype{R}{>{\raggedleft\arraybackslash}X}
\begin{document}

% \title{A Survey and Comparative Analysis of NLP-Based Deception Models Across Legal and General-Domain Texts}
% \title{A Comparative Survey of NLP Deception Detection in Legal vs General Domains}
% \title{ Legal Deception Detection with NLP: A Survey and Cross-Domain Comparative Analysis}
% \title{Benchmarking Deception Detection: A Cross-Domain Study of LLMs and Fine-tuned Models in Legal and General Contexts}
\title{Semantics of Subterfuge: Benchmarking Legal Deception Detection Against General-domain State-of-the-Art}


\author{Theekshana Samaradiwakara}
\email{theekshana.18@cse.mrt.ac.lk}
\orcid{0009-0007-4971-9613}
\author{Nisansa de Silva}
\email{NisansaDdS@cse.mrt.ac.lk}
\orcid{0000-0002-5361-4810}
\affiliation{%
 \institution{Department of Computer Science \& Engineering}
 \city{University of Moratuwa}
 \country{Sri Lanka}}

\author{George C. Lobb}
\email{Info@lobb.law}
\affiliation{%
  \institution{The Law Office of George C. Lobb}
  \city{Austin, Texas}
  \country{USA}
}

% \author{Anonymous Authors}
% \affiliation{
% \institution{Anonymous Affiliations}
% }

% TODO - Update


\begin{abstract} 
Deception detection has critical implications for legal proceedings, law enforcement, and online security. Although human judgment is limited in accuracy and scalability, Natural Language Processing (NLP) offers a data-driven alternative. We present a survey and comparative analysis of NLP-based Automatic Deception Detection (ADD) focusing on the legal domain, reviewing the evolution from feature-based machine learning to Large Language Model (LLM) approaches. We conduct a unified empirical evaluation across seven datasets (two legal, five general-domain), comparing six fine-tuned transformer models and seven LLMs under four prompting strategies. The results show strong domain sensitivity, with fine-tuned models excelling in data-rich general domains and few-shot LLMs remaining competitive in low-resource legal settings. Chain-of-Thought prompting often underperforms direct classification. These findings highlight the need for domain adaptation and interpretable systems in high-stakes legal contexts. \end{abstract}




% TODO Update

\maketitle

\section{Introduction}
% \textbf{TODO: Add more focus on legal domain in Introduction}
Deception detection is a critical challenge in various domains, including criminal investigations, legal proceedings, fraud prevention, and social media, with major implications for law enforcement and judicial systems. In legal contexts, the stakes are particularly high: undetected deception can obstruct justice, while false positives risk wrongful accusations and loss of public trust. Current practises are based on subjective judgment and resource-intensive manual analysis of statements and testimony. Automated deception detection systems could support decision-making by flagging suspicious statements while reducing the cognitive load on human analysts. However, the legal domain poses distinct challenges: limited training data due to privacy constraints, high costs of false positives, and strict requirements for interpretability and accountability.

\subsection{Scope and Contributions}
We present a structured review and comparative analysis of NLP-based deception detection with emphasis on legal and police contexts. We trace the evolution from traditional machine learning to modern LLM approaches, and conduct unified experiments across seven public datasets (two legal, five general-domain), comparing six fine-tuned transformer models (RoBERTa, BERT, DeBERTa, ALBERT, DistilBERT, T5) and seven LLMs (GPT-4o, LLaMA, Gemma2, Phi variants) under four prompting strategies. Our analysis reveals domain sensitivity patterns, prompting strategy effectiveness, temperature sensitivity, and practical limitations to guide future research toward effective and accountable ADD systems.

\section{Related Work}

\subsection{Evolution of NLP in Deception Detection}
Early deception research primarily focused on non-verbal signals. However, increasing evidence suggests that linguistic behaviour provides more reliable and scalable indicators of deceptive intent, enabling computational analysis without continuous human intervention. Empirical studies indicate that human judges perform only slightly above chance (54-60\% accuracy) in deception detection~\cite{li2020detection}, motivating the adoption of computational approaches that can identify subtle linguistic patterns across large text corpora. 

Text-based Automatic Deception Detection (ADD) has evolved through three main phases. Early methods relied on manually engineered linguistic and psychological features (Linguistic Inquiry and Word Count (LIWC) indicators, n-grams, POS (Part-of-Speech) tags) combined with classical machine learning classifiers (Support Vector Machines (SVM), Random Forests). Neural representation learning enabled models to capture semantic and contextual information directly from text, reducing dependence on handcrafted features. Most recently, large language models  (LLMs) have introduced knowledge-augmented approaches leveraging pre-training, prompt engineering, few-shot learning, and parameter-efficient fine-tuning for context-sensitive reasoning about deceptive language \cite{zhang2023towards}.

\subsection{Domain-Specific Applications: Legal and Policing} 
In legal contexts, deception detection supports trial transcript analysis, evidence validation, and credibility assessment, which raises ethical concerns around interpretability and fairness. Police-oriented research addresses complaint verification, misconduct detection, and false report identification. Table~\ref{tab:work_legal_police} summarises key work in both domains. Challenges include data scarcity, sensitivity, and high misclassification costs; thus, human oversight remains essential for responsible deployment.

\begin{table}[t]
  \caption{Work Related to the Legal and Police Domain}
  \label{tab:work_legal_police}
  % \small
  \resizebox{\columnwidth}{!}{%
  \begin{tabular}{p{.5cm} p{.5cm} p{6cm} p{4cm} p{4.5cm}}
    \toprule
    \textbf{Study} & \textbf{Year} & \textbf{Application} & \textbf{Related Data/Datasets} & \textbf{Model(s)} \\
    \midrule
    \cite{perez2015deception} & 2015 & Deception classification & RLTD dataset \cite{perez2015deception} & Decision Tree, Random Forest \\
    \midrule
    \cite{velutharambath2023unidecor} & 2023 & Cross-Corpus deception detection & RLTD dataset \cite{perez2015deception} & RoBERTa-base \\
    \midrule
    \cite{nguyen2024deception} & 2024 & Deception classification & RLTD dataset \cite{perez2015deception} & BiLSTM \\
    \midrule
    \cite{miah2025hidden} & 2025 & Direct label prediction, Post-hoc reasoning & RLTD dataset \cite{perez2015deception} &  LLaMA3.1-8B, Gemma2-9B, GPT-4o \\
    \midrule
    \cite{fornaciari2012use} & 2012 & Deception classification & DECOUR dataset \cite{fornaciari2012decour} & SVM \\
    \midrule
    \cite{fornaciari2013automatic} & 2013 & Deception classification & DECOUR dataset \cite{fornaciari2012decour} & SVM \\
    \midrule
    \cite{fornaciari2021bertective} & 2021 & Deception classification & DECOUR dataset \cite{fornaciari2012decour} & BERT + Transformers \\
    \midrule
    \cite{quijano2018applying} & 2018 & Detection of false robbery reports & Spanish police reports & SVM, Ridge LR \\
    \midrule
    \cite{chen2024can} & 2024 & Deception reasoning for criminal law documents  &  CAIL2018 \cite{xiao2018cail2018}, Synthetic dialogues & GLM-4-9B, GPT-3.5, Gemini-1.5-Pro, Qwen2-7B\\
    \bottomrule
  \end{tabular}
  }
\end{table}

\section{Comparative Analysis}
\subsection{Experimental Setup}
\subsubsection{DataSets}

Based on the existing work, we selected seven datasets: two legal domain datasets and five general domain datasets with the aim of evaluating cross-domain model behaviour. Table \ref{tab:dataset_statistics} summarises corpus statistics.

\begin{table}[t]
  \caption{Dataset Statistical Information. \texttt{SC}: average Sentence Count, \texttt{AT}: Average Token count, \texttt{VR}: Vocabulary Richness}
  \label{tab:dataset_statistics}
  \resizebox{\columnwidth}{!}{
  % \small
  \begin{tabular}{cccrrrrrrr}
    \toprule
    \textbf{Dataset} & \textbf{Domain} & \textbf{Data Type} & \textbf{Truthful} & \textbf{Deceptive} & \textbf{Total} & \textbf{SC} & \textbf{AT} & \textbf{VR} & \textbf{Vocab} \\
    \midrule
    RLTD \cite{perez2015deception} & legal & interview & 60 (49.6\%) & 61 (\textbf{50.4\%}) & 121 & 3.92 & 78.18 & \textbf{0.160} & 1,514 \\
    DECOUR \cite{fornaciari2012decour} & legal & interview & 1,202 (\textbf{56.0\%}) & 945 (44.0\%) & 2,147 & 1.02 & 14.97 & 0.092 & 2,956 \\
    OpSpam \cite{ott2011finding} & general & review & 800 (50.0\%) & 800 (50.0\%) & 1,600 & 9.54 & 167.73 & 0.039 & 10,497 \\
    cCult \cite{perez2014cross} & general & opinion & 606 (50.0\%) & 606 (50.0\%) & 1,212 & 4.41 & 77.04 & 0.064 & 6,003 \\
    DeRev2014 \cite{fornaciari2014identifying} & general & review & 118 (50.0\%) & 118 (50.0\%) & 236 & 6.69 & 141.96 & 0.144 & 4,839 \\
    Liar \cite{wang2017liar} & general & news & \textbf{7,134} (55.8\%) & \textbf{5,657} (44.2\%) & \textbf{12,791} & 1.17 & 20.21 & 0.059 & 15,186 \\
    FakeNewsNet \cite{shu2020fakenewsnet} & general & news & 211 (50.0\%) & 211 (50.0\%) & 422 & \textbf{27.01} & \textbf{712.19} & 0.052 & \textbf{15,534} \\
    \bottomrule
  \end{tabular}
  }
\end{table}

\paragraph{Real-Life Trial Deception (RLTD)~\cite{perez2015deception}} was created using the videos collected from public court trials in USA. The deception labels (truthful/deceptive) were determined by the trial outcomes. The videos were then transcribed via crowdsourcing to capture verbal information, and manually annotated for non-verbal cues such as facial displays and hand movements. For our experiments, we used the transcribed text without the non-verbal cues.

\paragraph{DECOUR~\cite{fornaciari2012decour}} includes Italian courtroom transcripts from 35 criminal hearings, comprising dialogues between interviewees and interviewers (judges, prosecutors, lawyers). Utterances were labelled True, False, or Uncertain. For our experiments, only True/False utterances were used to maintain binary classification consistency across datasets.

\paragraph{General Domain Datasets} \textbf{OpSpam~\cite{ott2011finding}} consists of hotel reviews collected from \textit{TripAdvisor} covering 20 Chicago hotels. Cross-cultural Deception Detection \textbf{(cCult)} \cite{perez2014cross} dataset was collected via crowdsourcing with truthful and deceptive short essays on opinions of three topics (Abortion, Death Penalty, Best Friend). \textbf{DeRev2014} \cite{fornaciari2014identifying} is a corpus of 236 book reviews, \textbf{Liar} \cite{wang2017liar} and \textbf{FakeNewsNet} \cite{shu2020fakenewsnet} are fake news datasets sourced from \textit{PolitiFact} and \textit{BuzzFeed}.

Table~\ref{tab:dataset_statistics} reports corpus sizes, label distributions, and linguistic characteristics. SC and token counts were computed using NLTK; vocabulary richness (VR) is defined as the type-token ratio, where lower values indicate more repetitive language.

\subsubsection{Data processing}
We applied unified 80:10:10 train-validation-test splits across all datasets, following original split ratios of Liar~\cite{wang2017liar}. However, its original six labels were mapped to binary: true/mostly-true/half-true $\to$ True and false/barely-true/pants-on-fire $\to$ False.

\subsubsection{Models}
We evaluated encoder-only transformers (RoBERTa, BERT, DeBERTa, ALBERT, DistilBERT) and T5-base via supervised fine-tuning, alongside seven LLMs: GPT-4o, GPT-4o-mini, LLaMA3-8B, LLaMA3.1-8B, Gemma2-9B, Phi-3-mini, and Phi-4, which have been used in recent work on deception detection \cite{benny2023knowledge, velutharambath2023unidecor, miah2025hidden, cui2025toward, aspromonte2025beyond, papantoniou2025evaluating}. Weighted F1 is used throughout to account for class imbalance.

\subsubsection{Experimental Setup}
Fine-tuning used: learning rate=2e-5, batch size=8, epochs=6, AdamW optimiser, on an NVIDIA RTX 3080 (16GB). Open-source LLMs were run locally via Ollama (v0.13.5) with 4-bit quantisation (Q4\_K\_M). OpenAI models were accessed via the Chat Completions API. All LLMs used temperature=0 for deterministic outputs unless stated otherwise.

\subsection{Prompting Strategies}
\subsubsection{Prompt Configurations} 
We evaluated five distinct experimental configurations to assess the impact of fine-tuning and prompting strategies: 
(1) \textbf{Supervised Fine-tuning:} BERT-based models and T5-base models were fine-tuned on training data using standard cross-entropy loss with the hyperparameters specified in Section 3.1.4.
(2) \textbf{Zero-shot Direct Classification:} LLMs received a task description and text input without examples:
(3) \textbf{Few-shot Direct Classification:} Four selected examples from the training set were prepended:
(4) \textbf{Zero-shot Chain-of-Thought (CoT):} Models were instructed to reason before classifying:
(5) \textbf{Few-shot Chain-of-Thought (CoT):} Combining examples with reasoning demonstrations (4 examples with step-by-step reasoning).

\subsubsection{Few-shot Example Selection} 

For all few-shot configurations, $k{=}4$ examples were selected dynamically per test instance using sentence embeddings (\texttt{all-MiniLM-L6-v2}), with label balancing enforced (2 truthful, 2 deceptive). \textbf{Top-K} selects the $k$ most similar examples to the query by cosine similarity. \textbf{High-Variance} uses a greedy procedure seeded by the most similar example, iteratively adding candidates that maximise pairwise similarity variance within the selected set, producing a linguistically diverse context.

\section{Results and Observations}

Fine-tuned transformers and LLMs were evaluated under consistent splits and prompts with no dataset-specific tuning. Full results are shown in Table~\ref{tab:dataset_f1_models_split}.

\begin{table*}
\caption{F1 Scores (\%) for Deception Detection Across Datasets and Models.
FT=Fine-tuned; ZS=Zero-shot; TK=Few-shot Top-K; HV=Few-shot High-Variance; CoT=Chain-of-Thought. 
Bold indicates best per dataset per section.}
\label{tab:dataset_f1_models_split}
%\small
\resizebox{\textwidth}{!}{
\begin{tabularx}{1.2\textwidth}{llRRRRRRR}
%\begin{tabular}{p{5cm} p{1.2cm} p{1.2cm} p{1.2cm} p{1.2cm} p{1.2cm} p{1.2cm} p{1.2cm}}
\toprule
{\footnotesize\textbf{Model}} & 
{\footnotesize\textbf{Strategy}} & 
{\footnotesize\textbf{RLTD \cite{perez2015deception}}} & 
{\footnotesize\textbf{DECOUR~\cite{fornaciari2012decour}}} & 
{\footnotesize\textbf{OpSpam~\cite{ott2011finding}}} & 
{\footnotesize\textbf{cCult \cite{perez2014cross}}} & 
{\footnotesize\textbf{DeRev2014~\cite{fornaciari2014identifying}}} & 
{\footnotesize\textbf{Liar~\cite{wang2017liar}}} & 
{\footnotesize\textbf{FakeNewsNet~\cite{shu2020fakenewsnet}}} \\
\midrule
\multicolumn{9}{l}{\textit{Transformer Fine-tuned}} \\
\midrule
RoBERTa-base & FT & 62.94 & 68.92 & 91.28 & 65.94 & \textbf{100.00} & 63.42 & 64.55 \\
BERT-base & FT & 48.11 & 68.31 & 90.02 & \textbf{69.74} & 96.00 & 63.71 & 57.46 \\
DeBERTa & FT & 77.53 & 70.80 & \textbf{92.52} & 68.96 & \textbf{100.00} & 62.28 & 62.93 \\
ALBERT & FT & 53.85 & 68.05 & 90.07 & 63.46 & 96.00 & 61.82 & 64.55 \\
DistilBERT & FT & 77.76 & \textbf{71.95} & 88.75 & 68.84 & \textbf{100.00} & 60.14 & 63.63 \\
T5-base & FT & 14.48 & 68.42 & 90.67 & 65.91 & 47.33 & 62.38 & \textbf{67.33} \\
\midrule
\multicolumn{9}{l}{\textit{LLM Direct Classification , Zero-shot}} \\
\midrule
Gemma2-9B  & ZS      & 24.90 & 38.77 & 53.01 & 55.13 & 54.56 & 56.88 & 67.11 \\
LLaMA 3-8B  & ZS     & 29.23 & 53.86 & 48.08 & 54.22 & 44.29 & 55.61 & 48.15 \\
LLaMA 3.1-8B  & ZS   & 54.95 & 51.43 & 45.07 & 53.41 & 37.10 & 58.03 & 36.34 \\
Phi-3-mini & ZS      & 14.48 & 26.56 & 36.91 & 40.01 & 35.58 & 25.69 & 49.12 \\
Phi-4 &  ZS          & 14.48 & 25.82 & 47.87 & 51.29 & 45.35 & 40.29 & 53.96 \\
GPT-4o &  ZS          & 41.59 & 40.52  & 59.63 & 53.97 & 55.86 & 60.19 & 65.92 \\
GPT-4o-mini &  ZS     & 52.20 & 45.67 & 55.93 & 55.97 & 49.60 & 53.95 & 59.11 \\
\midrule
\multicolumn{9}{l}{\textit{LLM Direct Classification , Few-shot (4-shot)}} \\
\midrule
Gemma2-9B   & TK / HV    & 54.95 / 70.38 & 55.92 / 57.47 & 46.99 / 51.91 & 49.33 / 51.75 & 58.00 / 71.08 & 59.20 / 59.94 & 66.67 / 62.97 \\
LLaMA 3-8B  & TK / HV    & 29.23 / 41.59 & 40.67 / 55.44 & 52.53 / 54.71 & 58.77 / 60.61 & 43.82 / 41.20 & 54.64 / 53.22 & 67.27 / 65.83 \\
LLaMA 3.1-8B & TK / HV   & 46.15 / 52.20 & 58.08 / 59.77 & 58.54 / 58.15 & 58.33 / 57.24 & 56.00 / 51.06 & 60.67 / 60.49 & 67.11 / 63.96 \\
Phi-3-mini  & TK / HV    & 56.64 / 65.64 & 53.38 / 48.34 & 39.58 / 40.52 & 37.99 / 34.21 & 29.33 / 29.33 & 59.02 / 58.94 & 56.25 / 60.04 \\
Phi-4   & TK / HV        & 14.48 / 29.23 & 30.53 / 36.35 & 42.71 / 42.53 & 50.23 / 52.71 & 57.83 / 69.20 & 42.60 / 40.08 & 60.00 / 63.96 \\
GPT-4o   & TK / HV       & \textbf{84.62} / \textbf{84.62} & 66.16 / 62.19 & 72.08 / 66.89 & 59.66 / 62.46 & 87.76 / 91.92 & 61.73 / 60.79 & 58.25 / 63.37 \\
GPT-4o-mini  & TK / HV   & 70.38 / 70.38 & 61.63 / 59.87 & 59.24 / 67.73 & 60.92 / 58.24 & 63.65 / 62.82 & 57.80 / 57.61 & 57.83 / 61.98 \\

\midrule
\multicolumn{9}{l}{\textit{LLM CoT , Zero-shot}} \\
\midrule
Gemma2-9B & ZS      & 61.54 & 46.50 & 63.31 & 53.34 & 63.88 & 50.91 & 53.33 \\
LLaMA 3-8B & ZS     & 75.88 & 51.86 & 49.99 & 53.49 & 58.00 & 55.94 & 62.97 \\
LLaMA 3.1-8B & ZS   & 41.59 & 53.84 & 50.64 & 53.62 & 64.00 & 54.55 & 57.40 \\
Phi-3-mini & ZS     & 53.85 & 55.23 & 38.63 & 47.20 & 31.14 & 60.57 & 62.97 \\
Phi-4 & ZS          & 56.64 & 39.78 & 29.61 & 39.07 & 31.14 & 55.00 & 19.67 \\
GPT-4o & ZS         & 71.65 & 46.71 & 49.60 & 51.26 & 54.56 & \textbf{64.26} & 55.06 \\
GPT-4o-mini & ZS    & 47.56 & 50.68 & 47.20 & 48.51 & 52.00 & 61.80 & 55.06 \\
\midrule
\multicolumn{9}{l}{\textit{LLM CoT , Few-shot (4-shot)}} \\
\midrule
Gemma2-9B   & TK / HV    & 54.95 / 62.94 & 57.85 / 58.69 & 52.47 / 56.04 & 54.70 / 49.85 & 72.00 / 71.08 & 57.02 / 56.66 & 56.23 / 53.33 \\
LLaMA 3-8B  & TK / HV    & 62.94 / 69.23 & 63.62 / 54.02 & 51.01 / 54.64 & 56.79 / 49.74 & 41.33 / 41.33 & 62.26 / 61.89 & 59.11 / 59.11 \\
LLaMA 3.1-8B & TK / HV   & 24.90 / 52.20 & 55.71 / 56.25 & 54.66 / 60.81 & 45.99 / 51.99 & 59.22 / 64.00 & 58.70 / 58.71 & 60.00 / 67.11 \\
Phi-3-mini  & TK / HV    & 56.64 / 56.64 & 51.78 / 49.95 & 33.23 / 34.85 & 43.84 / 36.74 & 29.33 / 29.33 & 60.11 / 59.88 & 54.15 / 60.04 \\
Phi-4  & TK / HV         & 56.64 / 56.64 & 44.36 / 45.45 & 29.61 / 29.61 & 26.71 / 28.48 & 31.14 / 31.14 & 54.12 / 53.98 & 32.75 / 26.50 \\
GPT-4o  & TK / HV        & 62.94 / 75.88 & 64.17 / 60.70 & 45.95 / 42.43 & 59.02 / 59.91 & 53.38 / 59.38 & 63.28 / 64.30 & 30.81 / 38.50 \\
GPT-4o-mini  & TK / HV   & 59.80 / 69.23 & 48.23 / 48.45 & 33.28 / 31.39 & 50.88 / 51.80 & 41.33 / 32.30 & 62.30 / 63.45 & 47.41 / 60.56 \\
\bottomrule
%\end{tabular}
\end{tabularx}
}
\end{table*}

\begin{table}
\caption{F1 Weighted Scores (\%) — 10-Fold CV vs Original (RLTD)}
\label{tab:rltd_cv_all}
\resizebox{\columnwidth}{!}{
% \small
\begin{tabular}{llrrrr}
\toprule
\textbf{Model} & \textbf{Setting} & \textbf{Original F1} & \textbf{CV F1 Weighted} & \textbf{Difference} & \textbf{std} \\
\midrule
\multicolumn{5}{l}{\textit{Transformer Fine-tune}} \\
\midrule
RoBERTa-base & Fine-tuned & 62.94 & 62.29 & -0.65 & 0.1057 \\
BERT-base    & Fine-tuned & 48.11 & 62.86 & +14.75 & 0.1180 \\
DeBERTa      & Fine-tuned & 77.53 & 61.46 & -16.07 & 0.0940\\
ALBERT       & Fine-tuned & 53.85 & 59.12 & +5.27 & 0.1281\\
DistilBERT   & Fine-tuned & 77.76 & 62.48 & -15.28 & 0.1954\\
T5-base      & Fine-tuned & 14.48 & 33.77 & +19.29 & 0.0131\\
\midrule
\multicolumn{5}{l}{\textit{LLM Direct Classification, Zero-shot}} \\
\midrule
Gemma2-9B    & Zero-shot & 24.90 & 48.14 & +23.24 & 0.0962\\
LLaMA 3.1-8B & Zero-shot & 54.95 & 52.69 & -2.26 & 0.0849\\
LLaMA 3-8B   & Zero-shot & 29.23 & 51.37 & +22.14 & 0.1660\\
Phi-3-mini   & Zero-shot & 14.48 & 35.39 & +20.91 & 0.0491\\
Phi-4        & Zero-shot & 14.48 & 39.93 & +25.45 & 0.0981\\
GPT-4o       & Zero-shot & 41.59 & 53.64 & +12.05 & 0.1254\\
GPT-4o-mini  & Zero-shot & 52.20 & 56.86 & +4.66 & 0.1114\\
\midrule
\multicolumn{5}{l}{\textit{LLM Direct Classification, Few-shot (4-shot, Top-K Selection)}} \\
\midrule
Gemma2-9B    & Few-shot Top-K & 54.95 & 68.21 & +13.26 & 0.1122\\
LLaMA 3.1-8B & Few-shot Top-K & 46.15 & 59.80 & +13.65 & 0.1541\\
LLaMA 3-8B   & Few-shot Top-K & 29.23 & 50.89 & +21.66 & 0.1412\\
Phi-3-mini   & Few-shot Top-K & 56.64 & 44.12 & -12.52 & 0.1296\\
Phi-4        & Few-shot Top-K & 14.48 & 44.11 & +29.63 & 0.1155\\
GPT-4o       & Few-shot Top-K & 84.62 & 78.14 & -6.48 & 0.0533 \\
GPT-4o-mini  & Few-shot Top-K & 70.38 & 71.61 & +1.23 & 0.1147 \\
\midrule
\multicolumn{5}{l}{\textit{LLM Direct Classification, Few-shot (4-shot, High Variance Selection)}} \\
\midrule
Gemma2-9B    & Few-shot High-Var & 70.38 & 63.72 & -6.66 & 0.0809\\
LLaMA 3.1-8B & Few-shot High-Var & 52.20 & 65.19 & +12.99 & 0.1437 \\
LLaMA 3-8B   & Few-shot High-Var & 41.59 & 48.38 & +6.79 & 0.1163\\
Phi-3-mini   & Few-shot High-Var & 65.64 & 48.56 & -17.08 & 0.1383\\
Phi-4        & Few-shot High-Var & 29.23 & 44.76 & +15.53 & 0.1146\\
GPT-4o       & Few-shot High-Var & 84.62 & 75.01 & -9.61 & 0.1188 \\
GPT-4o-mini  & Few-shot High-Var & 70.38 & 63.75 & -6.63 & 0.0885 \\
\bottomrule
\end{tabular}
}
\end{table}

\begin{table}
\caption{F1 Weighted Scores (\%) — LLM Direct Classification, 10-Fold CV Across Temperatures (RLTD)}
\label{tab:rltd_llm_temps}
% \small
\resizebox{\columnwidth}{!}{%
\begin{tabular}{llrrrrrrrr}
\toprule
\textbf{Model} & \textbf{Setting} & \textbf{t=0.0} & \textbf{t=0.3} & \textbf{t=0.5} & \textbf{t=0.7} & \textbf{t=1.0} & \textbf{Mean} & \textbf{std} \\
\midrule
\multicolumn{9}{l}{\textit{Zero-shot}} \\
\midrule
Gemma2-9B    & Zero-shot & 48.14 & 47.42 & 48.48 & 53.01 & 49.50 & 49.31 & 1.97 \\
LLaMA 3.1-8B & Zero-shot & 52.69 & 51.15 & 57.46 & 50.76 & 55.57 & 53.53 & 2.59 \\
LLaMA 3-8B   & Zero-shot & 51.37 & 55.77 & 52.46 & 54.16 & 51.40 & 53.03 & 1.70 \\
Phi-3-mini   & Zero-shot & 35.39 & 33.77 & 33.77 & 33.77 & 33.77 & 34.09 & 0.65 \\
Phi-4        & Zero-shot & 39.93 & 39.93 & 39.68 & 41.56 & 44.47 & 41.11 & 1.81 \\
\midrule
\multicolumn{9}{l}{\textit{Few-shot (4-shot, Top-K Selection)}} \\
\midrule
Gemma2-9B    & Few-shot Top-K & 68.21 & 67.26 & 69.08 & 67.30 & 70.02 & 68.37 & 1.06 \\
LLaMA 3.1-8B & Few-shot Top-K & 59.80 & 63.15 & 58.59 & 63.66 & 55.51 & 60.14 & 3.01 \\
LLaMA 3-8B   & Few-shot Top-K & 50.89 & 50.89 & 50.89 & 47.60 & 49.27 & 49.91 & 1.31 \\
Phi-3-mini   & Few-shot Top-K & 44.12 & 46.06 & 46.16 & 47.29 & 46.24 & 45.97 & 1.03 \\
Phi-4        & Few-shot Top-K & 44.11 & 42.48 & 44.47 & 44.81 & 45.73 & 44.32 & 1.07 \\
\midrule
\multicolumn{9}{l}{\textit{Few-shot (4-shot, High Variance Selection)}} \\
\midrule
Gemma2-9B    & Few-shot High-Var & 63.72 & 64.65 & 65.40 & 66.26 & 62.04 & 64.41 & 1.45 \\
LLaMA 3.1-8B & Few-shot High-Var & 65.19 & 65.37 & 62.27 & 53.71 & 63.69 & 62.05 & 4.32 \\
LLaMA 3-8B   & Few-shot High-Var & 48.38 & 49.08 & 49.67 & 49.67 & 49.46 & 49.25 & 0.49 \\
Phi-3-mini   & Few-shot High-Var & 48.56 & 47.57 & 50.16 & 49.70 & 48.70 & 48.94 & 0.91 \\
Phi-4        & Few-shot High-Var & 44.76 & 48.01 & 41.82 & 49.30 & 50.45 & 46.87 & 3.16 \\
\bottomrule
\end{tabular}
}
\end{table}

\paragraph{Fine-tuned Models vs. Zero-shot LLMs}
Fine-tuned transformer models consistently outperformed zero-shot LLMs on datasets with sufficient data. DeBERTa achieves F1 = 92.52\% on OpSpam versus GPT-4o's 59.63\%; multiple fine-tuned models reach 100\% on DeRev2014 hough this reflects dataset artefacts (see Section~\ref{sec:disc}). However, on the small RLTD dataset (121 samples), few-shot GPT-4o (F1=84.62\%) matches or exceeds fine-tuned models, suggesting pre-trained priors compensate when labelled data is scarce.

\begin{figure}[h]
  \centering
  \includegraphics[width=\linewidth]{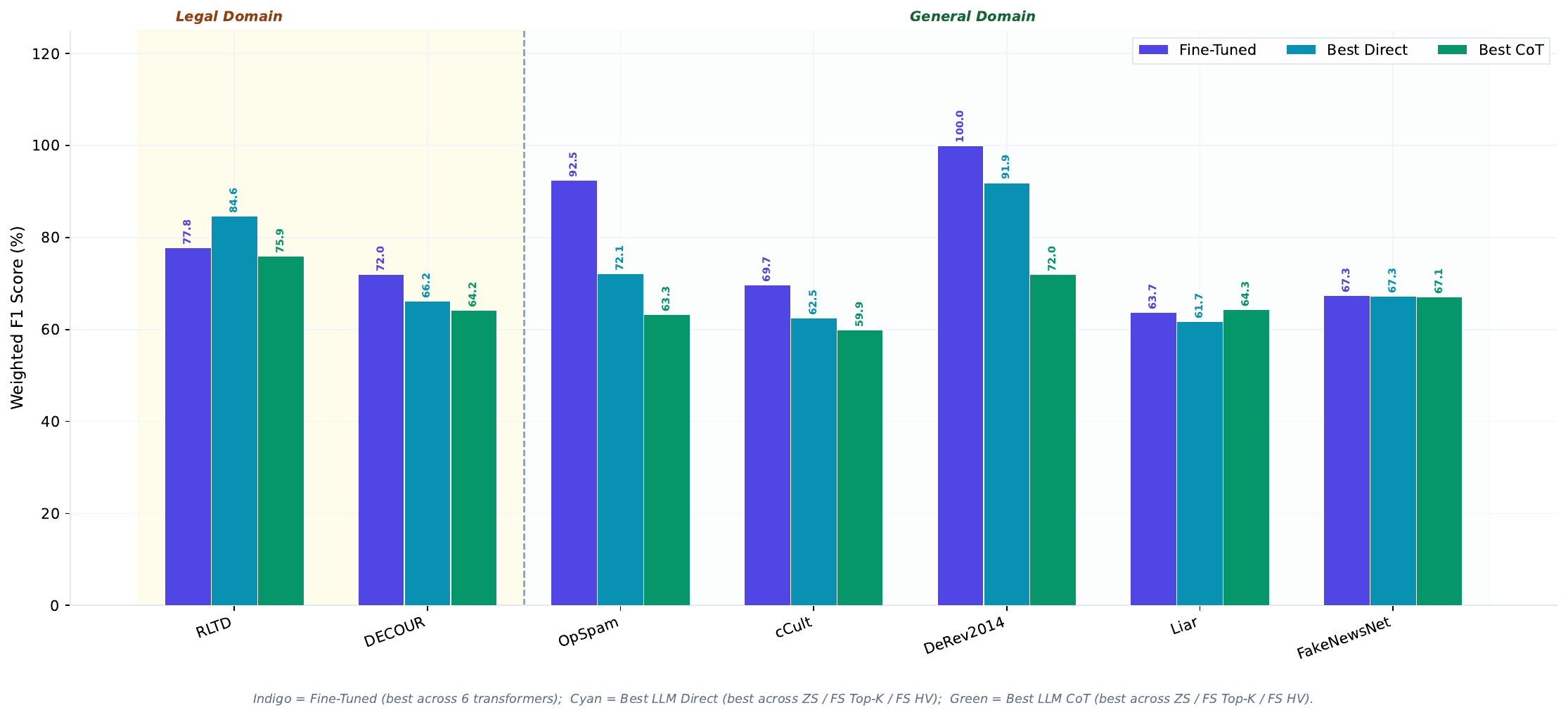}
  \caption{Best F1 per method per dataset.}
  \label{fig:overview_best_per_method}
\end{figure}

\paragraph{Domain Sensitivity}
Legal datasets yield lower, more variable performance: RLTD spans 14.48--84.62\% across models; DECOUR peaks at 71.95\% (DistilBERT). Review datasets are most tractable (OpSpam: 92.52\%, DeRev2014: 100\%), while news datasets show intermediate difficulty, with GPT-4o achieving 65.92\% zero-shot on FakeNewsNet, likely benefiting from world knowledge. Figure~\ref{fig:overview_best_per_method} summarises the best achievable F1 per method per dataset, illustrating the consistent advantage of fine-tuning over LLM-based approaches across six of the seven datasets, with the exception of RLTD where few-shot LLMs prevail.

% \begin{figure}[h]
%   \centering
%   \large
%   \includegraphics[width=1\linewidth]{fig_e.png}
%   \caption{Domain-level Performance Trends}
%   \Description{\textbf{Domain-level Performance Trends}}
% \end{figure}

% \begin{figure*}[h]
%   \centering
%   \large
%   \includegraphics[width=0.7\linewidth]{fig_f.png}
%   \caption{Best Approach for Each Domain}
%   \Description{\textbf{Best Approach for Each Domain}}
% \end{figure*}

\begin{figure}[h]
  \centering
  \includegraphics[width=\linewidth]{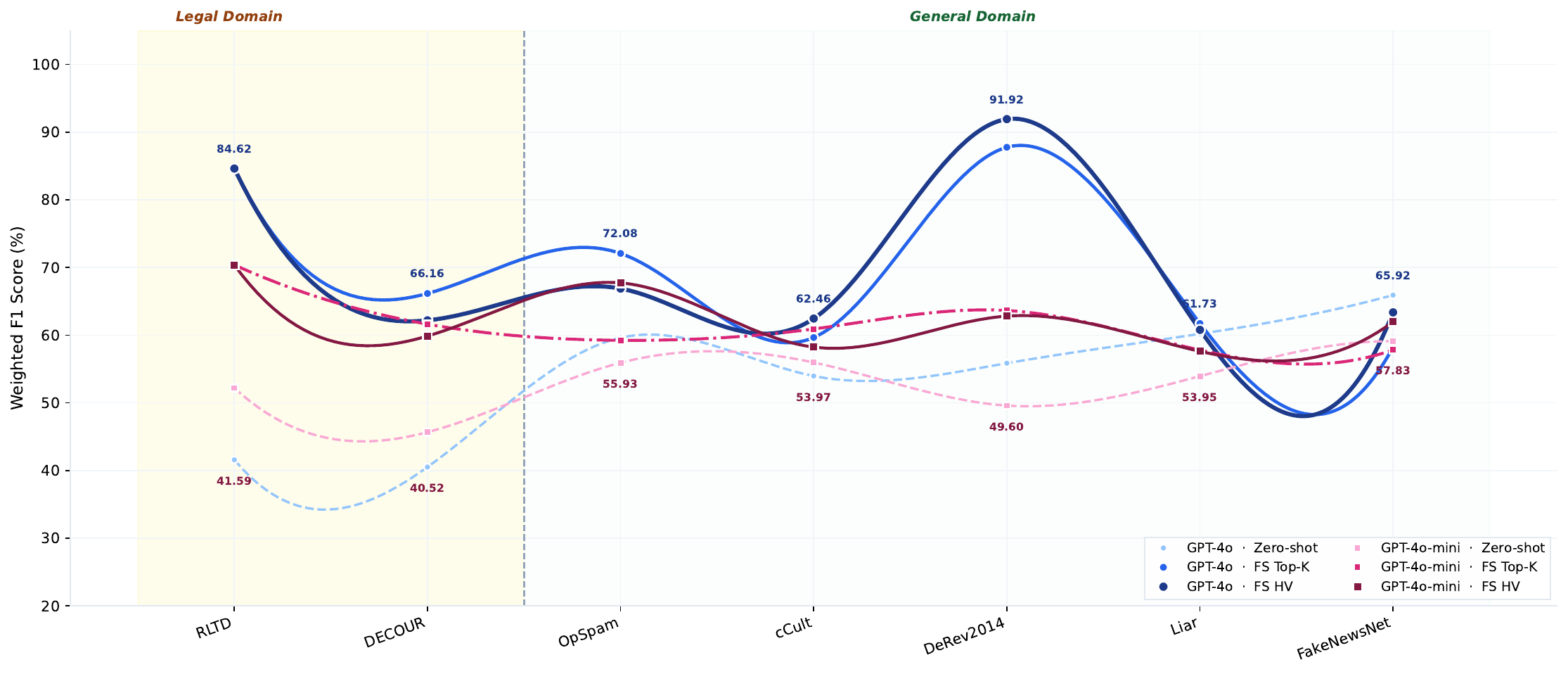}
  \caption{GPT-4o \& GPT-4o-mini: Zero-Shot vs Few-Shot (Direct Classification)}
  \label{fig:gpt4o_direct_lines}
\end{figure}

\paragraph{Impact of Few-shot Learning}
Few-shot prompting yields inconsistent gains. GPT-4o improves substantially on RLTD (41.59\% to 84.62\%) and DECOUR (40.52\% to 66.16\%) with 4-shot Top-K. Conversely, LLaMA models degrade on several datasets under few-shot prompting. High-variance example selection occasionally outperforms Top-K (e.g., GPT-4o on DeRev2014, High-Variance (91.92\%) outperforms Top-K (87.76\%)) but shows no consistent advantage, confirming sensitivity to example choice. Figure~\ref{fig:gpt4o_direct_lines} traces GPT-4o and GPT-4o-mini across all three direct-classification shot variants and all seven datasets; annotated values show the per-dataset maximum (dark blue) and minimum (dark red) across all six variants.

\paragraph{Chain-of-Thought Reasoning}
CoT shows inconsistent effects: GPT-4o zero-shot CoT improves over zero-shot direct on RLTD (41.59\% to 71.65\%), yet 4-shot direct (84.62\%) outperforms 4-shot CoT (62.94\%). CoT consistently underperforms direct classification on OpSpam.  Dataset-specific benefits appear on DeRev2014 (LLaMA3-8B CoT: 58.00\% vs.\ direct: 44.29\%). These findings align with~\cite{miah2025hidden}, confirming reasoning steps do not universally benefit deception detection.

\paragraph{Model Size and Architecture Effects}
Larger models do not guarantee better performance: GPT-4o outperforms GPT-4o-mini but margins are modest. Among fine-tuned models, DeBERTa and DistilBERT show strongest overall performance; ALBERT lags despite architectural similarity to BERT. T5-base underperforms on small datasets (RLTD: 14.48\%) but remains competitive on larger corpora.

\paragraph{Stability Analysis (RLTD)}
Given RLTD's small size, we performed 10-fold cross-validation (Table~\ref{tab:rltd_cv_all}). High standard deviations confirm substantial instability: BERT improves from 48.11\% to 62.86\% CV mean, while DeBERTa drops from 77.53\% to 61.46\%, indicating overfitting on single splits. Among LLMs, GPT-4o with Top-K selection achieves the highest CV mean (78.14\%, std=0.053), the most stable strong result across all configurations.

\paragraph{Temperature Sensitivity (RLTD)}
Table~\ref{tab:rltd_llm_temps} shows LLM performance across temperatures (0.0--1.0) on RLTD. Most models exhibit low sensitivity: Gemma2-9B few-shot Top-K varies only 1.06 std across temperatures. LLaMA3.1-8B shows higher variance under few-shot High-Variance selection (std=4.32), suggesting interaction between example diversity and output stochasticity. Phi-3-mini produces near-constant output across temperatures in zero-shot settings (std=0.65 across folds), suggesting degenerate behaviour.

\section{Discussion}
\label{sec:disc}

% \textbf{TODO: complete the discussion with our results}
After achieving the 100\%  F1 score for DeRev2014 across multiple models, we looked for existing work where similar results are reported. \citet{papantoniou2022deception} report identical results and identifies lexical leakage: words such as \textit{thriller} appear exclusively in deceptive samples, while \textit{Stephen} appears only in truthful ones, arising from the use of different books per class. This constitutes a dataset quality artefact rather than genuine generalisation, underscoring the need for carefully constructed benchmarks.

Our results confirm strong domain dependence with limited cross-domain transfer. Fine-tuned models optimised on review data cannot be deployed in legal contexts without substantial degradation, consistent with prior findings on domain-specific deceptive cues~\cite{sarzynska2023truth,loconte2025detecting}. This motivates domain-adaptive training or multi-domain fine-tuning pipelines.

A clear trade-off emerges between fine-tuning and prompting. Datasets with sufficient samples and stylistic regularity (OpSpam, DECOUR) benefit from fine-tuning. On small legal datasets (RLTD: 121 samples), few-shot GPT-4o (84.62\%) avoids the overfitting risks of fine-tuning. For legal practitioners with limited annotations, few-shot LLMs offer a practical entry point; organisations with annotation capacity should invest in fine-tuned models for cost-effective deployment.

Counterintuitively, explicit reasoning via CoT often degrades performance relative to direct classification. This suggests that deception detection relies more on pattern recognition than on systematic logical inference, contrasting with CoT's effectiveness on tasks such as mathematical reasoning and highlighting the need for task-specific prompt engineering.

% \textbf{Implications for Legal Deployment}

\section{Conclusion}
We present a survey and comparative evaluation of NLP-based deception detection with emphasis on legal and law enforcement contexts. Experiments across seven datasets and thirteen models under multiple prompting strategies yield several actionable findings. We demonstrate that model performance is highly domain-dependent necessitating domain-specific adaptation. Fine-tuning outperforms prompting given sufficient data, but few-shot GPT-4o achieves competitive performance (84.62\% F1) on small legal datasets. Chain-of-thought prompting shows inconsistent effects, often underperforming direct classification, challenging assumptions on universal utility of reasoning. No single model dominates across all datasets; optimal selection depends on domain, dataset size, and linguistic complexity.
For legal practitioners, current ADD systems should serve as decision-support tools rather than autonomous mechanisms. Critical gaps remain in interpretability, bias mitigation, and adversarial robustness. Future work should prioritise domain adaptation, interpretable fine-tuning, and the construction of large-scale legal deception corpora with rich metadata to advance trustworthy deployment.

% \begin{acks}

% \end{acks}

\bibliographystyle{ACM-Reference-Format}
\bibliography{base}

\appendix

\end{document}